\documentclass[runningheads]{llncs}
\usepackage[T1]{fontenc}
\usepackage{graphicx}
\usepackage{amsmath}
\begin{document}
\title{AnoGAN for Tabular Data: A Novel Approach to Anomaly Detection}
\author{Pavan Reddy\inst{1} \and
Aditya Singh\inst{1} }
\footnotetext[1]{Both authors contributed equally to this research.}
\authorrunning{P. Reddy and A. Singh}
\institute{George Washington University, Washington, DC 20052, USA\\
\email{\{pavan.reddy,adityasingh\}@gwu.edu}
}
\maketitle              % typeset the header of the contribution
\begin{abstract}
Anomaly detection, a critical facet in data analysis, involves identifying patterns that deviate from expected behavior. This research addresses the complexities inherent in anomaly detection, exploring challenges and adapting to sophisticated malicious activities. With applications spanning cybersecurity, healthcare, finance, and surveillance, anomalies often signify critical information or potential threats. Inspired by the success of Anomaly Generative Adversarial Network (AnoGAN) in image domains, our research extends its principles to tabular data. Our contributions include adapting AnoGAN's principles to a new domain and promising advancements in detecting previously undetectable anomalies. This paper delves into the multifaceted nature of anomaly detection, considering the dynamic evolution of normal behavior, context-dependent anomaly definitions, and data-related challenges like noise and imbalances.

\end{abstract}
\section{Introduction}
Anomaly detection identifies unexpected patterns in data across various domains, often referred to as anomalies, outliers, or contaminants. It finds applications in cybersecurity, safety-critical system monitoring, fraud detection in finance and healthcare, and military surveillance for non-traditional enemy activities.

Anomaly detection, distinct from noise accommodation and removal \cite{chandola2009anomaly}, addresses unwanted noise in data. Noise, defined as irrelevant data phenomena hindering interpretation, requires elimination before analysis. Conversely, noise accommodation shields statistical model estimates from outlier impacts.

Anomaly detection plays a vital role across various domains by uncovering crucial insights from data anomalies. An unusual network traffic pattern \cite{ahmed2015novel} may indicate a security breach, while abnormal MRI scans \cite{wang2020brain} could signal the presence of tumors. Anomalies in aviation sensors \cite{basora2019recent} may highlight potential aircraft component issues, and deviations in credit card transactions often signify fraudulent activity. Anomaly detection finds practical applications in manufacturing, finance, and medical imaging, relying on models to identify abnormal patterns amidst regular data. Despite extensive research, managing complex, high-dimensional data remains a challenge. Various communities have developed specialized anomaly detection techniques tailored to specific domains.

Generative Adversarial Networks (GANs), introduced by Ian Goodfellow \cite{goodfellow2014generative} and colleagues, have emerged as a powerful modeling approach for handling high-dimensional data. Anomaly Generative Adversarial Network (AnoGAN) integrates traditional anomaly detection methods with GAN architecture, enabling it to generate data while simultaneously learning typical data properties for anomaly detection. 

Our main contributions in this paper can be summarized as follows:

\begin{itemize}
    \item How do different reconstruction errors affect the performance of GAN for anomaly detection in tabular data, and what thresholds or criteria can be established for effective detection?
    \item How does the performance of a GAN-based anomaly detection model with the optimal reconstruction error, trained on single-class data, compare to traditional anomaly detection methods in terms of accuracy and efficiency?
\end{itemize}

% \begin{figure}[!b]
%   \centering
%   \includegraphics[width=\columnwidth]{images/flow_chart_anomaly_detection.jpeg}
  
%   \caption{ Key components associated with an anomaly detection technique \cite{chandola2009anomaly}}
%   \label{fig:largegrid}
% \end{figure}

\subsection{Challenges in Anomaly Detection}
Identifying anomalies in datasets presents several challenges. Firstly, distinguishing abnormal patterns from normal ones requires algorithms that minimize false positives while effectively detecting anomalies. Defining "normal" behavior is complex and evolves over time, necessitating adaptable baseline models. Anomaly detection must also adapt to recognize malicious activities acing as normal behavior, balancing adaptability and resilience. Furthermore, normal behavior changes dynamically, requiring systems that learn and update accordingly. Context-dependent anomaly definitions add another layer of complexity, demanding models that contextualize data based on specific circumstances. Data-related challenges, including handling noisy or imbalanced datasets, addressing blind spots, and implementing preprocessing techniques, further complicate anomaly detection. Finally, managing the general complexity inherent in anomaly detection, such as balancing sensitivity and specificity and adapting to changing conditions, requires expertise in machine learning, data analysis, and domain knowledge.

\subsection{Overview of GANs}

The introduction of Generative Adversarial Networks (GANs) (Figure 2) by Ian Goodfellow and associates \cite{goodfellow2014generative} offers a strong modeling approach for addressing the problem of high-dimensional data. Two adversarial networks, a generator (G) and a discriminator (D) are involved in conventional GANs. While D learns to distinguish between actual data and samples provided by G, G is in charge of modeling the data by learning a mapping from latent random variables z (derived from Gaussian or uniform distributions) to the data space. GANs are seeing increasing use in speech and medical imaging, where they have shown empirical success as natural image models. 

% \begin{figure}[!h]
%   \centering
%   \includegraphics[width=\columnwidth]{images/gan_gen_dis.png}
  
%   \caption{GANs, or Generative Adversarial Networks, are intricate deep neural network structures consisting of two networks, namely the generator and discriminator. These networks work in opposition to each other, which is why they are called "adversarial." The generator accepts random numbers as input and produces an image. This generated image is then presented to the discriminator, along with a continuous stream of images sourced from the genuine, ground-truth dataset. \cite{article}}
%   \label{fig:largegrid}
% \end{figure}
\begin{figure}[!h]
  \centering
  \includegraphics[width=0.6\columnwidth]{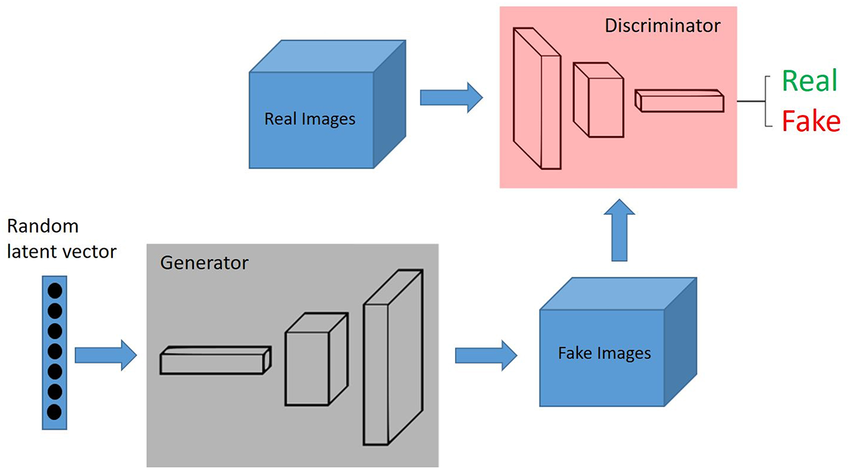}
  
  \caption{GANs, or Generative Adversarial Networks, are intricate deep neural network structures consisting of two networks, namely the generator and discriminator. These networks work in opposition to each other, which is why they are called "adversarial." The generator accepts random numbers as input and produces an image. This generated image is then presented to the discriminator, along with a continuous stream of images sourced from the genuine, ground-truth dataset. \cite{article}}
  \label{fig:largegrid}
\end{figure}

Anomaly Generative Adversarial Network (Figure 3), or AnoGAN \cite{schlegl2017unsupervised}, is a cutting-edge anomaly detection method that blends features of conventional anomaly detection approaches with generative adversarial networks (GANs). The GAN architecture is expanded to include anomaly detection with AnoGAN, which enables the model to produce data while concurrently learning the properties of typical data to detect anomalies.

\begin{figure}[!t]
  \centering
  \includegraphics[width=0.6\columnwidth]{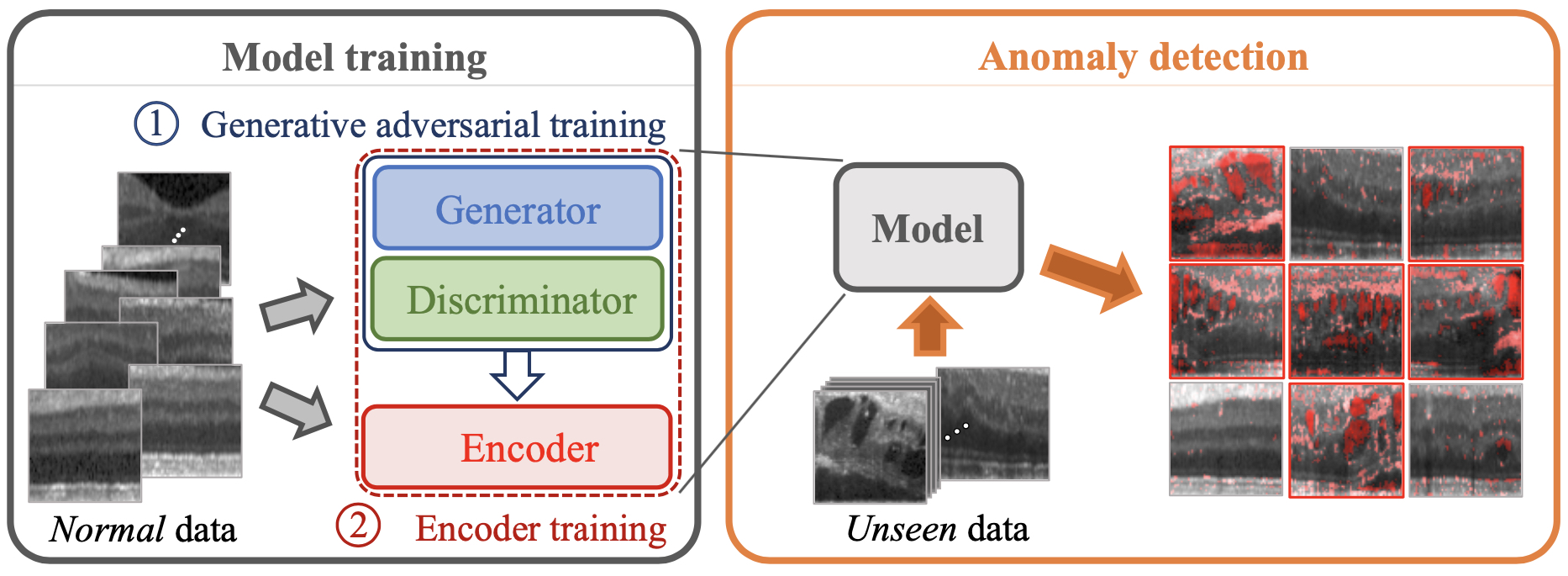}
  
  \caption{Framework for identifying anomalies. This framework involves two main stages of model training: generative adversarial training, which results in a trained generator and discriminator, and encoder training, which yields a trained encoder. Both of these training phases are executed using normal or "healthy" data. Subsequently, the framework is used for anomaly detection, where it is applied to both unseen healthy cases and anomalous data. \cite{schlegl2017unsupervised}}
  \label{fig:largegrid}
\end{figure}

AnoGAN's versatility enables its application across diverse domains for anomaly detection. Particularly in medical imaging, finance, and cybersecurity, it excels in identifying irregular patterns. In finance, it detects fraud by recognizing unusual transaction behaviors, while in cybersecurity, it spots anomalies in network traffic. Its adaptability to time series data facilitates predictive maintenance, anticipating equipment breakdowns. In text analysis, it identifies abnormal language patterns. Overall, AnoGAN offers a flexible solution for anomaly detection in various industries, from healthcare to industrial process monitoring.

\section{Related Work}

\begin{figure*}[b]  % Use [t] to place the image at the top of the page
  \centering
  \includegraphics[width=\textwidth]{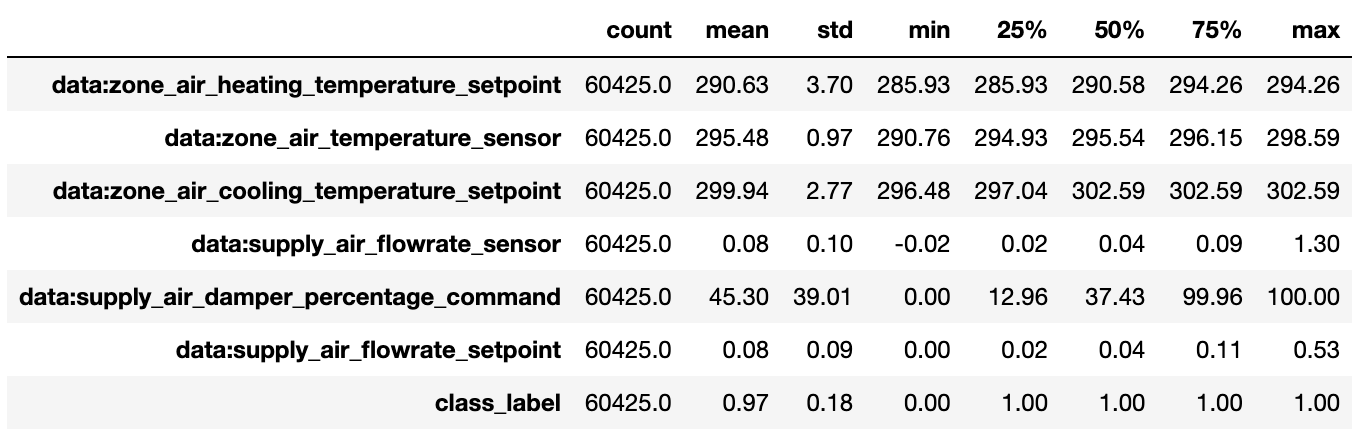}
  \caption{Statistical values of the data collected from the Google campus for the Variable Air Volume devices \cite{sipple2020interpretable}}
  \label{fig:your-image}
\end{figure*}

In the realm of deep learning-based anomaly detection, various approaches have surfaced over recent years. Researchers have \cite{chalapathy2019deep} comprehensively reviewed deep learning techniques for anomaly detection, emphasizing the promise and adaptability of these models across diverse data types. Their work echoes the sentiment of Zhou and Paffenroth \cite{zhou2017anomaly}, who proposed an anomaly detection method utilizing autoencoders, this method employs neural network architectures to reconstruct input data. Anomalies are identified when the reconstruction error surpasses a predefined threshold, enabling effective detection of deviations from normal patterns. They demonstrated their efficacy in capturing non-linear transformations and achieving impressive detection rates. Further, Ruff et al. \cite{pmlr-v80-ruff18a} extended the applicability of one-class classification using deep neural networks, shedding light on the strength of deep architectures in isolating normal from anomalous patterns. Their approach extends the scope of one-class classification through the integration of deep neural networks. The model is trained on normal instances, enhancing its ability to discern anomalies by learning complex patterns inherent in the normal data distribution. In the context of Generative Adversarial Networks (GANs), Zenati \cite{zenati2018adversarially} introduced an adversarially-trained one-class classifier, which proved pivotal in benchmark datasets. They introduced an adversarial element to one-class classification, this method employs a classifier trained to distinguish between normal and anomalous samples. Adversarial training enhances the model's robustness, making it more adept at discerning subtle anomalies. In another critical development, researchers \cite{de2019batch} explored the time series implications in anomaly detection, emphasizing the temporal dynamics of data. This method addresses anomalies by considering temporal dependencies. It accounts for patterns evolving over time, enabling the identification of deviations from expected temporal sequences. Moreover, there has been a growing interest \cite{zhang2020cluster} into hybrid models, merging traditional statistical methods with deep learning, establishing a bridge between classical and contemporary methodologies.

\section{Methodology}
The methodology is executed systematically, beginning with rigorous data collection and preprocessing to ensure the quality of the dataset. Our methodology tackles the issue of prediction randomness in CTGAN models, attributed to the Gumbel Softmax activation, which is essential for ensuring stable training. Instead of eliminating the Gumbel Softmax, we mitigate this randomness through the implementation of a hard Gumbel Softmax approach, enhancing the precision and reliability of the model's outputs. AnoGAN methodology is then applied for anomaly detection, involving the optimization of noise vectors through backpropagation using Mean Squared Error (MSE) loss. This process results in an anomaly score that reflects disparities between synthetic and original samples. Further refinement includes the determination of an optimal threshold through Receiver Operating Characteristic (ROC) analysis, enhancing the discriminative capacity of the framework. The investigation of individual feature differences between normal and synthetic samples concludes this methodological phase, offering detailed insights into the detected anomalies within structured datasets.

\subsection{Dataset}

The Smart Buildings Anomaly Detection dataset \cite{sipple2020interpretable}, covers 14 days from October 8 to 21, 2019, at a Google campus in California. It includes 60,425 (Figure 4) observations from 15 Variable Air Volume (VAV) devices, with 1,921 instances (3.2\%) showing anomalies. These devices, designed to regulate air temperature, have two operational modes: a stricter comfort mode on weekdays (6:00 am to 10:00 pm) and a more relaxed eco mode during off-hours. The data is enriched with temporal markers like the day and hour.

\subsection{Data preprocessing}

In the initial phase of our methodology, the raw dataset is subjected to a thorough preprocessing pipeline to enhance its suitability for subsequent analysis. The dataset, once loaded undergoes a series of transformations for optimal utility. Initially, irrelevant features, namely 'dow' (day of the week) and 'hod' (hour of the day), are removed to streamline the dataset.

Further refinement includes categorizing the data into specific classes, with anomalies identified by the class label '0' being segregated into an exclusive dataframe for testing. Concurrently, data instances classified as normal are compiled into a distinct dataframe, as our training exclusively focuses on normal data.

To standardize and normalize the dataset, the Min-Max scaling technique is applied, rescaling feature values to a specified range (-1 to 1). This transformation is particularly valuable for maintaining the integrity of the anomaly and normal data distributions while preparing them for integration into the subsequent anomaly detection framework. \\

The multimodal input data undergoes processing via a Gaussian Mixture Model to derive multiple unimodal representations. This methodology enhances the stability of training Generative Adversarial Networks (GANs) by facilitating the learning of a broader array of simpler distributions, as opposed to a singular complex distribution. \\

$p(x) = \sum_{i=1}^{M} \pi_i \mathcal{N}(x | \mu_i, \Sigma_i)\rightarrow$ GMM Transformation\\\\
where \(p(x)\) is the probability density function of the data \(x\) modeled as a mixture of \(M\) Gaussian distributions, \(\pi_i\) are the mixing coefficients, \(\mu_i\) and \(\Sigma_i\) are the mean and covariance of the \(i\)-th Gaussian component, respectively.

\subsection{CT-GAN Implementation and Randomness Handling}

In our research methodology, the utilization of the original CT-GAN (Conditional Tabular Generative Adversarial Network) (Figure 5) \cite{xu2019modeling} implementation plays a pivotal role in generating synthetic samples for anomaly detection. CT-GAN is a specialized GAN variant designed for tabular data, aiming to faithfully reproduce the statistical characteristics of the given dataset. However, the original CT-GAN implementation introduces a layer of unpredictability during the testing phase, manifested in the randomness inherent in the predictions generated by the generator. To address this challenge and enhance the stability of the generated samples, we employ a modified softmax gumbel activation. This modification is instrumental in mitigating the unpredictable nature of the test predictions, ensuring a more consistent and reliable generation of synthetic samples. The application of the softmax gumbel activation contributes to the robustness and effectiveness of our anomaly detection framework by providing a more controlled and deterministic generation process for the synthetic data. Figure 6 shows the KDE for both the real and generated data after applying CT-GAN. \\
\\
% Original Gumbel Softmax\\
% $G_i = -\log(-\log(U_i))$ where $U_i$ is $Uniform(0,1)\rightarrow$ Gumbel Noise \\\\
% $h_i = \frac{l_i + G_i}{\tau}\rightarrow$ Temperature Scaling Logits\\\\
% $y_i = \frac{\exp(h_i)}{\sum_{j} \exp(h_j)}\rightarrow$-> Softmax on logits
% \\\\
% Hard Gumbel Softmax\\\\
% $y_i = \frac{\exp(l_i / \tau)}{\sum_{j} \exp(l_j / \tau)}\rightarrow$ Temperature Scaling and softmax\\\\
% $y_{\text{hard}, i} = \begin{cases} 1 & \text{if } i = \arg\max_j y_j \\ 0 & \text{otherwise} \end{cases}$ Selecting the max component\\\\
% $y = \text{stop\_gradient}(y_{\text{hard}} - y) + y \rightarrow$ Straight through Estimator\\\\

\textbf{Original Gumbel Softmax}
\begin{align*}
G_i & = -\log(-\log(U_i)) && \text{where } U_i \text{ is } Uniform(0,1) \text{ and } G_i \text{ is Gumbel Noise} \\
h_i & = \frac{l_i + G_i}{\tau} && \text{where } \tau \text{ is Temperature Scaling Constant and }  l_i \text{ is logits} \\
y_i & = \frac{\exp(h_i)}{\sum_{j} \exp(h_j)} && \text{Softmax on noised and scaled logits}
\end{align*} \\

\textbf{Hard Gumbel Softmax}
\begin{align*}
y_i & = \frac{\exp(l_i / \tau)}{\sum_{j} \exp(l_j / \tau)} && \text{Temperature Scaling and softmax on logits} \\
y_{\text{hard}, i} & = \begin{cases} 1 & \text{if } i = \arg\max_j y_j \\ 0 & \text{otherwise} \end{cases} && \text{Selecting the max component} \\
y & = \text{stop\_gradient}(y_{\text{hard}} - y) + y && \text{Straight through Estimator}
\end{align*}

\subsection{Optimizing Noise Vector and Anomaly Scoring}

\begin{figure*}[t]  % Use [t] to place the image at the top of the page
  \centering
  \includegraphics[width=\textwidth]{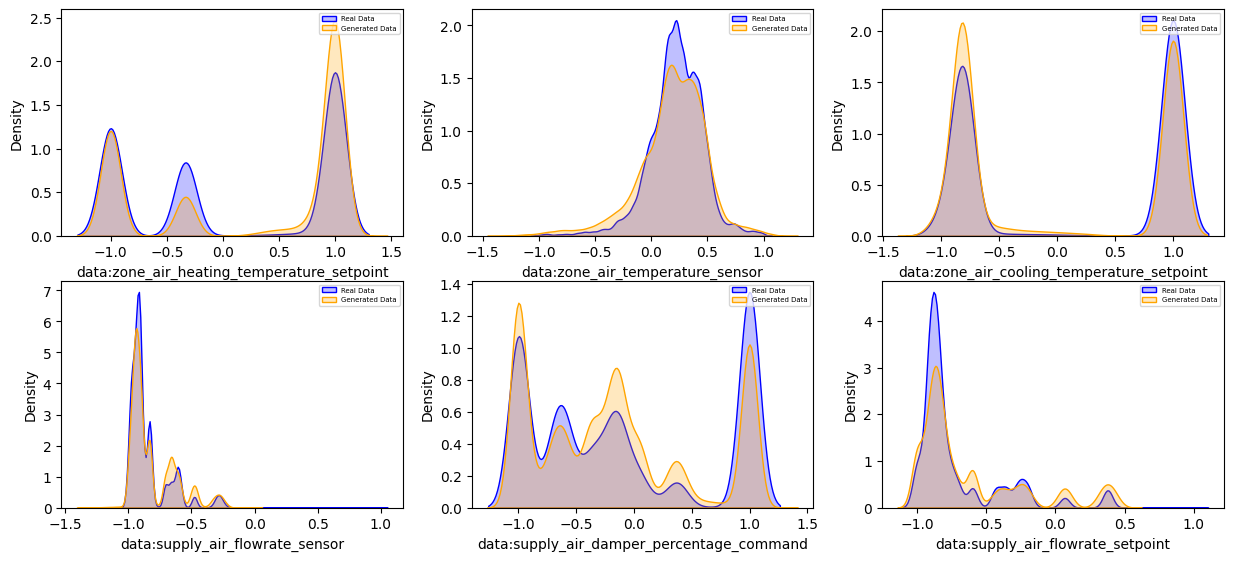}
  \caption{The figure provides a visual representation of the Kernel Density Estimation (KDE) for both the real and generated datasets. The KDE serves as a smoothed probability density function, offering insights into the underlying distribution of the data. Specifically, the KDE for the real data showcases the probability density across different values, illustrating the distribution's characteristics and patterns. On the other hand, the KDE for the generated data, produced by the Generator component of the Generative Adversarial Network (GAN), offers a comparative view. This comparison enables an assessment of how well the generator has learned to replicate the statistical properties of the real data. By visually inspecting the KDE curves, one can discern the fidelity of the generated data distribution in relation to the authentic dataset, providing a valuable tool for evaluating the performance and quality of the GAN's generative capabilities.}
  \label{fig:your-image}
\end{figure*}
In this phase of our methodology, the focus is on refining the generated synthetic samples to closely mirror the characteristics of the original data. For a given sample, we initiate the optimization process of the noise vector, a critical parameter in the generative process. The objective is to adjust the noise vector such that the generator produces synthetic samples that are the most similar to the corresponding original samples.

To quantify the dissimilarity between the synthetic and original samples, we employ the Mean Squared Error (MSE) loss during the backpropagation process to optimize the latent vector. We tried several other loss-measuring functions but MSE worked the best for the data. The MSE loss serves as a measure of the average squared differences between corresponding elements of the synthetic and original samples. This penalizes larger differences and allows us to capture the nuances and intricacies of the data distribution, facilitating a more precise evaluation of the generative process.

The computed MSE loss between the original sample and generated sample serves as the anomaly score, representing the magnitude of deviation between the two samples. This score becomes a crucial metric for distinguishing normal from anomalous samples. Leveraging this anomaly score, we establish a threshold that optimally discriminates between normal and anomalous instances, determined through Area Under Curve-Receiver Operating Characteristic (AUC-ROC) analysis. This meticulous threshold determination enhances the discriminative power of our anomaly detection framework.

Furthermore, delving into individual feature differences between normal and synthetic samples provides insights into the specific aspects contributing to detected anomalies. By scrutinizing these differences, we gain a nuanced understanding of the reasons behind the anomalies, allowing for more informed and targeted interventions in subsequent stages of analysis. This comprehensive approach to anomaly scoring and threshold determination forms a robust foundation for the effectiveness of our anomaly detection methodology.

\section{Results}

In this section, we present a comprehensive evaluation of our anomaly detection framework based on AnoGAN, comparing its performance against two alternative methodologies, namely One-Class Support Vector Machine (OCSVM) and k-Nearest Neighbors (KNN). The choice of OCSVM and KNN as comparative methods reflects their widespread usage and effectiveness in anomaly detection tasks. Table 1 captures the accuracy of all the methods implemented. Through a rigorous analysis of the results obtained from these three approaches, we aim to elucidate the strengths and limitations of our AnoGAN-based framework in detecting anomalies within structured datasets. 
\begin{table}[h]
\centering
\begin{tabular}{lccc}
\hline
\textbf{Methods} & \textbf{AUC-ROC} \\
\hline
OCSVM & 55.6\% \\
KNN & 50\% \\
AnoGAN & 72\% \\
\hline
\end{tabular}
\caption{Comparison of Anomaly Detection Methods based on Accuracy.}
\label{tab:accuracy-comparison}
\end{table}

The progression of Generative Adversarial Network (GAN) losses across epochs is a key aspect of our findings. As the GAN undergoes training, the Generator Loss, representing the ability to generate realistic data, exhibits notable fluctuations. Initially high, this loss gradually diminishes as the Generator refines its capacity to produce more authentic samples. Simultaneously, the Discriminator Loss, indicating the Discriminator's accuracy in distinguishing real from generated data, follows a similar trajectory. After a certain point the loss starts to increase but our implementation stops the progression with early stopping. The delicate interplay between these losses is pivotal for achieving equilibrium in GAN training and generating high quality samples.

\begin{figure}[!h]
  \centering
  \includegraphics[width=0.6\columnwidth]{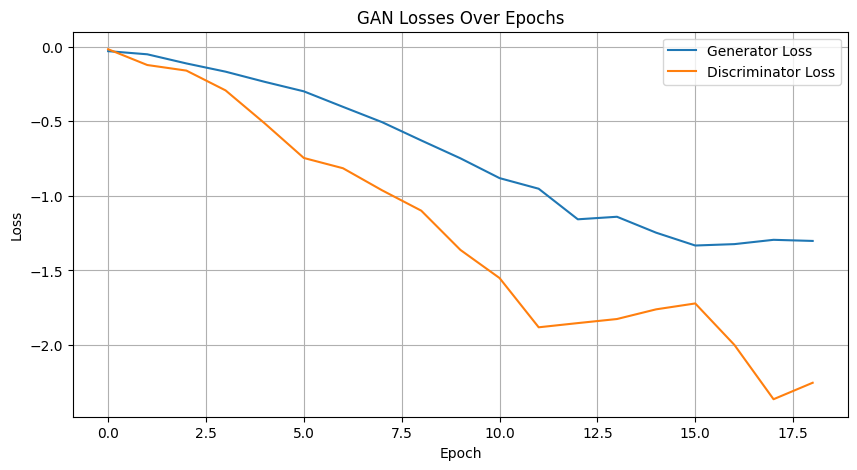}
  
  \caption{The figure illustrates the progression of Generator and Discriminator losses over training epochs in the Generative Adversarial Network (GAN), showcasing how the model refines generative and discriminative capabilities during learning.}
  \label{fig:largegrid}
\end{figure}

\begin{figure}[!t]
  \centering
  \includegraphics[width=0.6\columnwidth]{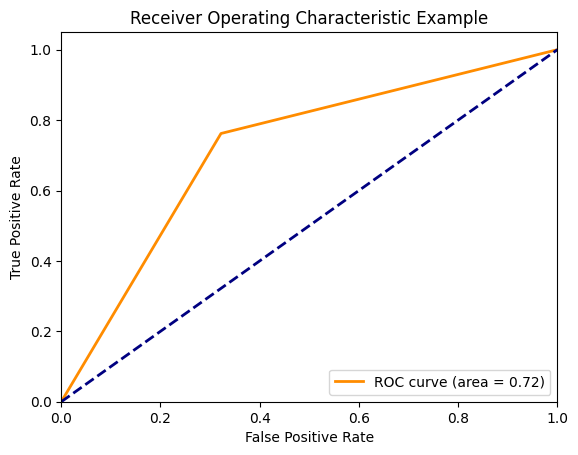}
  
  \caption{The ROC curve visually represents the Receiver Operating Characteristic for the Generative Adversarial Network (GAN), offering insights into its discriminative performance and the trade-off between true positive and false positive rates}
  \label{fig:largegrid}
\end{figure}

Our innovative approach involves applying AnoGAN to synthetic data generated by CTGAN and normal data samples, achieving a robust anomaly detection accuracy of approximately 72\%. Figure 7 shows the loss over epoch for training and Figure 8 shows the ROC curve for our method. Extended training substantially improves accuracy, reaching around 80\%. This outperforms traditional methods like KNN and OCSVM, where accuracy is comparable to random assignment at 50\%. The superior performance underscores the efficacy of our methodology in detecting anomalies with higher precision and reliability.

Our methodology demonstrates notable efficacy in scenarios characterized by infrequent data anomalies. The system exhibits exceptional performance, achieving an accuracy exceeding 85\% when tested on datasets featuring a mere 50 anomaly samples. The observed variance in accuracy, contingent upon the anomaly count, can be attributed to the inherent challenge of discerning the underlying generative function for anomalies. This task is comparatively more straightforward with a smaller anomaly set, where distinct patterns are easily recognizable, as opposed to the heightened complexity associated with larger anomaly datasets.

\section{Conclusion and Future work}
Our meticulously executed methodology encompasses systematic data collection and preprocessing, ensuring dataset quality. Incorporating the original CT-GAN implementation, we address inherent randomness in test predictions through softmax gumbel activation, enhancing the stability of generated samples. Leveraging AnoGAN for anomaly detection, our approach optimizes noise vectors using Mean Squared Error (MSE) loss, resulting in a nuanced anomaly score reflective of synthetic-original disparities. Refinement involves ROC analysis for threshold determination, enhancing the framework's discriminative capacity.

Looking ahead, our research opens avenues for several promising directions. Firstly, exploring the integration of domain-specific knowledge into the anomaly detection process can enhance the model's interpretability and performance. Additionally, investigating the adaptability of our methodology to incorporate categorical variables will broaden its applicability across a wide range of domains. Further research into refining the threshold determination process and extending the framework to handle dynamic datasets with evolving patterns could deepen its practical applicability. The exploration of ensemble techniques and hybrid models, combining the strengths of different anomaly detection methods, presents another intriguing avenue for future investigation. Continuous refinement and adaptation of our methodology will be essential for addressing evolving challenges in anomaly detection across various real-world scenarios.

\end{document}